\documentclass{article}

\PassOptionsToPackage{numbers, compress}{natbib}

\usepackage[final]{neurips_2023}

\usepackage[utf8]{inputenc} 
\usepackage[T1]{fontenc}    
\usepackage[hidelinks]{hyperref}       
\usepackage{url}            
\usepackage{booktabs}       
\usepackage{amsfonts}       
\usepackage{nicefrac}       
\usepackage{microtype}      
\usepackage{xcolor}         
\usepackage{xspace}         
\usepackage{enumitem}       
\usepackage{amsmath}
\usepackage{amssymb}
\usepackage{multirow}
\usepackage{graphicx}
\usepackage[nohyperlinks]{acronym}
\makeatletter
\DeclareRobustCommand\onedot{\futurelet\@let@token\@onedot}
\def\@onedot{\ifx\@let@token.\else.\null\fi\xspace}

\def\eg{\emph{e.g}\onedot} 
\def\ie{\emph{i.e}\onedot}

\def\etal{\emph{et al}\onedot}
\makeatother

\DeclareMathOperator*{\argmax}{arg \ max}

\definecolor{myred}{RGB}{222,119,174}
\definecolor{mygreen}{RGB}{77,146,33}
\definecolor{cnn}{RGB}{227, 164, 0} 
\definecolor{convstem}{RGB}{0, 133, 255} 
\definecolor{linear}{RGB}{19, 139, 0} 

\acrodef{tsr}[TSR]{table structure recognition}
\acrodef{teds}[TEDS]{tree-edit-distance-based similarity}
\acrodef{vlp}[VLP]{vision-language pretraining}
\acrodef{cnn}[CNN]{convolutional neural network}
\acrodef{pp}[pp]{percentage points}
\acrodef{sota}[SOTA]{state-of-the-art}
\acrodef{vit}[ViT]{vision Transformer}
\acrodef{relu}[ReLU]{rectified linear unit}
\acrodef{gelu}[GELU]{gaussian error linear unit}
\acrodef{nlp}[NLP]{natural language processing}
\acrodef{vilt}[ViLT]{vision-and-language transformer}
\acrodef{ssl}[SSL]{self-supervised learning}
\acrodef{rf}[RF]{receptive field}
\acrodef{iou}[IoU]{intersection over union}
\acrodef{mac}[MAC]{Multiply-Add Operations per Second}
\acrodef{edd}[EDD]{encoder-dual-decoder}
\acrodef{pdf}[PDF]{portable document format}
\title{High-Performance Transformers for Table Structure Recognition Need Early Convolutions}

\author{
  ShengYun Peng$^1$
  \quad
  Seongmin Lee$^1$
  \quad
  Xiaojing Wang$^2$ 
  \quad
  Rajarajeswari Balasubramaniyan$^2$
  \And
  Duen Horng Chau$^1$ \\
  \AND
  $^1$Georgia Institute of Technology
  \quad
  $^2$ADP, Inc. \\
  \texttt{\{speng65, seongmin, polo\}@gatech.edu} \\
  \texttt{\{xiaojing.wang, raji.balasubramaniyan\}@adp.com }
}

\begin{document}
\maketitle

\begin{abstract}

\Acf{tsr} aims to convert tabular images into a machine-readable format, where a visual encoder extracts image features and a textual decoder generates table-representing tokens.
Existing approaches use classic \ac{cnn} backbones for the visual encoder and transformers for the textual decoder. 
However, this hybrid \ac{cnn}-Transformer architecture introduces a complex visual encoder that accounts for nearly half of the total model parameters, markedly reduces both training and inference speed, and hinders the potential for self-supervised learning in \ac{tsr}.
In this work, we design a lightweight visual encoder for \ac{tsr} without sacrificing expressive power.
We discover that a convolutional stem can match classic \ac{cnn} backbone performance, with a much simpler model.
The convolutional stem strikes an optimal balance between two crucial factors for high-performance \ac{tsr}: a higher \ac{rf} ratio and a longer sequence length.
This allows it to ``see'' an appropriate portion of the table and ``store'' the complex table structure within sufficient context length for the subsequent transformer. 
We conducted reproducible ablation studies and open-sourced our code at \href{https://github.com/poloclub/tsr-convstem}{https://github.com/poloclub/tsr-convstem} 
to enhance transparency, inspire innovations, and facilitate fair comparisons in our domain as tables are a promising modality for representation learning.
\end{abstract}
\section{Introduction}
\label{sec:introduction}

\Acf{tsr} aims to extract both the structure and cell data of a tabular image into a machine-readable format~\cite{zhong2020image, nassar2022tableformer, huang2023improving}. 
This task is inherently an image-to-text generation problem, where a visual encoder extracts image features, and a textual decoder generates tokens representing the table, typically in HTML~\cite{li2019tablebank} or LaTeX symbols~\cite{ye2021pingan}.
In the existing literature, the visual encoder often employs classic \ac{cnn} backbones, \eg, ResNets and their variants~\cite{he2016deep}, and the textural decoder consists of a stack of transformer encoders and decoders~\cite{vaswani2017attention}. 
However, this hybrid \ac{cnn}-transformer architecture introduces a complex visual encoder that takes up almost half of the total model parameters and significantly reduces both training and inference speed~\cite{lu2019vilbert, chen2020uniter, huang2020pixel}.
Dosovitskiy, \etal~\cite{dosovitskiy2020image} compared the vanilla \ac{vit}, which used a simple linear projection, with the ``hybrid \ac{vit}'' that used 40 convolution layers (most of a ResNet-50) and found that both models performed similarly for image classification tasks. 
Furthermore, linear projection has proven to be a powerful visual processor, enabling \ac{ssl} in related domains, \eg, document image classification and layout analysis tasks~\cite{li2022dit, bao2021beit}. 
This leads us to ponder: 
How to simplify the visual encoder for \ac{tsr} while reducing computational costs without sacrificing performance?
Can we simply employ the aforementioned linear projection?

\begin{figure}[t]
\centering
\includegraphics[width=1\linewidth]{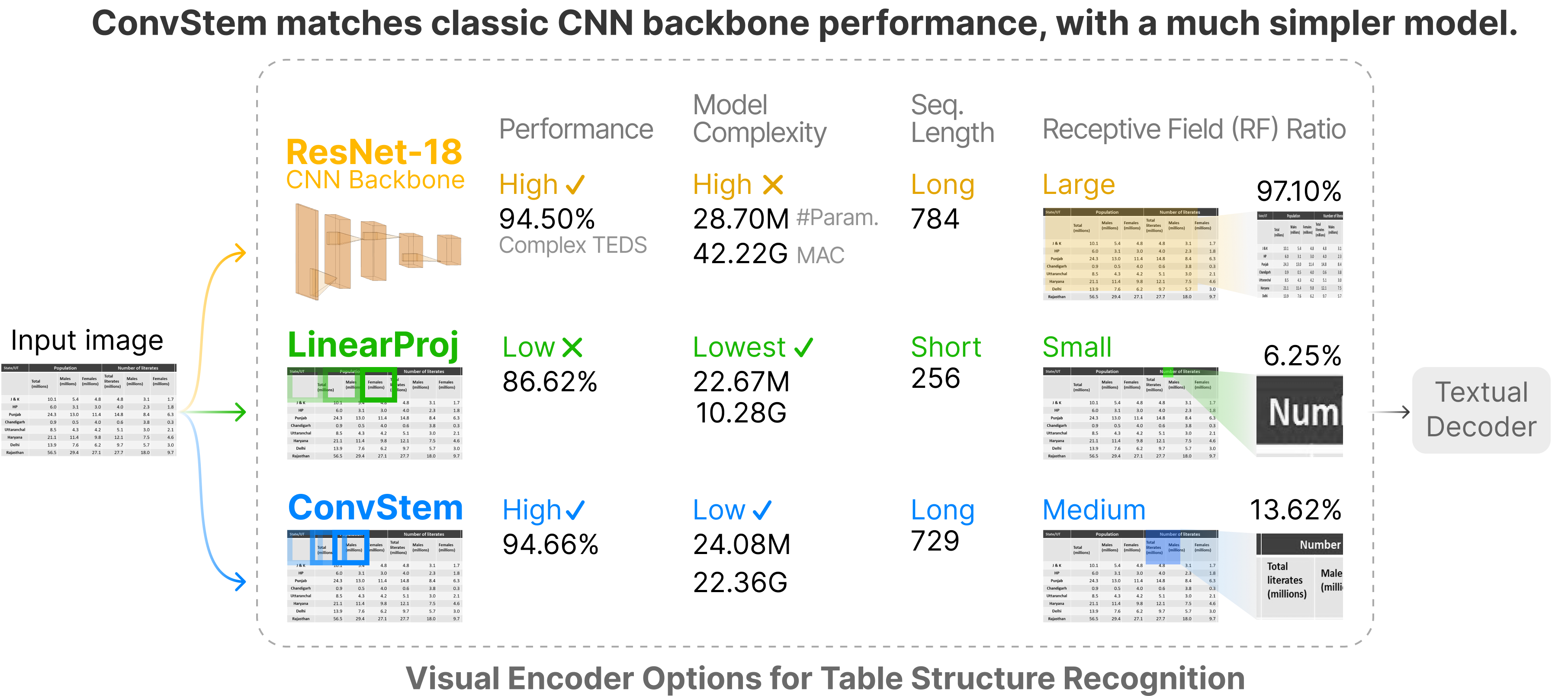}
\caption{
Using convolutional stem (ConvStem) in a visual encoder for table structure recognition (TSR) achieves performance comparable to that of a CNN backbone while significantly reducing model complexity.
The \ac{cnn} backbone is performant with large \ac{rf} but exhibits high model complexity. 
Linear projection is the simplest but suffers in terms of performance due to limited \ac{rf} and sequence length. 
In contrast, ConvStem strikes an optimal balance between two crucial factors for high-performance \ac{tsr}: a higher \textit{\acf{rf} ratio} and a longer \textit{sequence length}.
We illustrate each visual encoder option's \ac{rf} (zoomed in) and compute its \ac{rf} ratio. 
Using the image features extracted from the visual encoder, a textual decoder then generates tokens representing the table.
}
\label{fig:pipeline}
\end{figure}

We address the above research questions and make the following major contributions (Fig.~\ref{fig:pipeline}):
\begin{enumerate}[leftmargin=*,topsep=0pt]
\itemsep-0.1em 
\item \textbf{We discover that a convolutional stem can match classic \ac{cnn} backbone performance, with a much simpler model} (Fig.~\ref{fig:pipeline}: \textcolor{convstem}{bottom}).
This discovery stems from our motivation to design a lightweight visual encoder for \ac{tsr} without compromising on expressive power.
To achieve this, we began by replacing the \ac{cnn} backbone with \ac{vit}'s linear projection.
However, this substitution led to a noticeable $\sim 6$ \ac{pp} decrease in overall accuracy and $\sim 8$ \ac{pp} for complex tables.
The major difference between the \ac{cnn} backbone and the linear projection lies in the total number of convolution layers.
A standard ResNet-18 has 17 convolution layers, whereas the linear projection only has one. 
Evidence shows that incorporating a few early convolutions is crucial to balance inductive biases and the representation learning ability of transformers~\cite{xiao2021early}. 
We therefore reintroduce a few convolution layers by constructing the convolutional stem (Sec.~\ref{sec:visual}). 
This addition proves highly effective, bridging the performance gap between the linear projection and the \ac{cnn} backbone while still achieving low model complexity. 

\item \textbf{The convolutional stem strikes an optimal balance between two crucial factors for high-performance \ac{tsr}: a higher \acf{rf} ratio and a longer sequence length.}
The \textit{\ac{rf} ratio}, defined as the ratio of the \ac{rf} in the input to the size of the entire input, reflects how much of an input impacts the visual encoder's output.
As illustrated in Fig.~\ref{fig:pipeline}, a \textcolor{linear}{small} \ac{rf} ratio provides minimal structural information, while a \textcolor{convstem}{medium-sized} one offers ample context and distinguishable features.
\textit{Sequence length} refers to the transformer's input length, and longer context during training typically yields higher-quality models~\cite{tay2020long}. 
Consequently, the performance of linear projection is capped since the \textit{\ac{rf} ratio} and the \textit{sequence length} are inversely correlated, \ie, an increase in the \ac{rf} ratio means a larger patch size, resulting in a shorter sequence.  
In contrast, the convolutional stem independently balances these two factors, increasing the \ac{rf} ratio while maintaining the sequence length. 
This enables it to ``see'' an appropriate portion of the table and ``store'' the complex table structure within sufficient context length for the subsequent transformer. 
Additionally, with fewer convolutional layers than a typical \ac{cnn} backbone, the convolutional stem significantly reduces model complexity.

\item \textbf{Reproducible research and open-source code.}
We provide all the details regarding training, validation, and testing, which include model architecture configurations, model complexities, dataset information, evaluation metrics, training optimizer, learning rate, and ablation studies. 
Our work is open source and publicly available at  \href{https://github.com/poloclub/tsr-convstem}{https://github.com/poloclub/tsr-convstem}.
We believe that reproducible research and open-source code enhance transparency, inspire \ac{sota} innovations, and 
facilitate fair comparisons in our domain as tables are a promising modality for representation learning.
\end{enumerate}

\section{Related Work}
\label{sec:related_work}

\textbf{\ac{tsr} based on image-to-text generation.}
This method treats the table structure as a sequence and adopts an end-to-end image-to-text paradigm. 
Deng, \etal~\cite{deng2019challenges} employed a hybrid CNN-LSTM architecture to generate the LaTeX code of the table. 
Zhong, \etal~\cite{zhong2020image} introduced an \ac{edd} architecture in which two RNN-based decoders were responsible for logical and cell content, respectively. 
Both TableFormer~\cite{nassar2022tableformer} and TableMaster~\cite{ye2021pingan} enhanced the \ac{edd} decoder with a transformer decoder and included a regression decoder to predict the bounding box instead of the content.
VAST~\cite{huang2023improving} took a different approach by modeling the bounding box coordinates as a language sequence and proposed an auxiliary visual alignment loss to ensure that the logical representation of the non-empty cells contains more local visual details.
Our goal differs from existing approaches as we focused on designing a lightweight visual encoder and conducting a comprehensive comparison of three different types of visual encoders through ablation studies. 

\textbf{Applications of lightweight visual encoders}
While the transformer has become the solid mainstream for \ac{nlp} tasks, its applications in computer vision remained limited until the advent of \ac{vit}~\cite{dosovitskiy2020image}. 
For visual tasks, attention mechanisms were either applied in conjunction with \acp{cnn}~\cite{wang2018non} or used to replace components within \acp{cnn}~\cite{ramachandran2019stand}.
\Ac{vit} demonstrated that this reliance on \acp{cnn} is unnecessary and transformers can directly process sequences of image patches through linear projection. 
This same simplification also occurred in the \ac{vlp} domain, where early models used severely slow region selection~\cite{lu2019vilbert} or grid features~\cite{huang2020pixel}.
\Ac{vilt}~\cite{kim2021vilt} adopted the same linear projection from \ac{vit} to minimize overhead in the visual encoder. 
The convolutional stem~\cite{xiao2021early}, another lightweight visual encoder, reintroduced minimal convolutions to enhance optimization stability and improve peak performance and robustness~\cite{singh2023revisiting, peng2023robust}. 
It was initially introduced to replace the linear projection in \acp{vit}, serving as the earliest stage of input image processing~\cite{xiao2021early}. 
Lightweight visual encoders are not only used in supervised learning described above but also in \ac{ssl}. 
BEiT~\cite{bao2021beit} designed a masked image modeling task to pretrain vision transformers. 
Specifically, each image has two views: image patches from linear projection and visual tokens.
The pretraining objective is to recover the original visual tokens based on the corrupted image patches. 
Inspired by BEiT, DiT~\cite{li2022dit} proposed a self-supervised pre-trained document image Transformer model, which leveraged large-scale unlabeled document images for pre-training.
Hence, the lightweight visual encoder is a powerful input image processor.
In this work, we explore these lightweight visual encoders in \ac{tsr} for the first time.

\section{Discovering Visual Encoder Impact on TSR Architecture}
\label{sec:methodology}

\subsection{Overview of TSR Pipeline}
The goal of \ac{tsr} is to translate the input tabular image $I$ into a machine-readable sequence $T$. 
Specifically, $T$ includes table structure $T_s$ defined by HTML table tags and table content $T_c$ defined by standard \ac{nlp} vocabulary. 
The prediction of $T_c$ is triggered when a non-empty table cell is encountered in $T_s$, \eg, \texttt{<td>} for a single cell or \texttt{>} for a spanning cell. 
If the corresponding \ac{pdf} is provided, a cell coordinate decoder will predict the cell location and extract $T_c$ directly from the \ac{pdf}, which is also triggered by a non-empty table cell. 
Thus, accurate structure prediction is a bottleneck that affects the performance of the downstream cell data recognition. 
Our model focuses on the table structure prediction and comprises two modules: visual encoder and textual decoder. 
The visual encoder extracts image features and the textual decoder generates HTML table tags based on the image features. 
Sec.~\ref{sec:visual} and \ref{sec:structure} introduce the visual encoder and textual decoder and Sec.~\ref{sec:loss} presents the training loss function. 

\subsection{Comparing Visual Encoder Options} 
\label{sec:visual}
Given an input tabular image of size $(H, W)$, the visual encoder extracts the features required for downstream transformers. 
In this section, we compare three types of visual encoders: \ac{cnn} backbones, linear projection, and convolutional stem. 

\textbf{\ac{cnn} backbone.} 
Among the architectures used in current \ac{tsr} research, off-the-shelf \ac{cnn} backbones, especially ResNets and ResNet variants~\cite{he2016deep} are the most widely employed. 
EDD~\cite{zhong2020image} explored five different ResNet-18 variants, TableFormer~\cite{nassar2022tableformer} used a ResNet-18 with an additional adaptive pooling layer, and VAST~\cite{huang2023improving} modified a ResNet-31 equipped with multi-aspect global content attention~\cite{lu2021master}. 
We use models from Torchvision libraries~\cite{paszke2017automatic} and evaluate ResNet-18, ResNet-34, and ResNet-50. 
The penultimate pooling and final linear layers are removed and the output is the feature map from the last convolution layer.
All ResNet-18, ResNet-34, and ResNet-50 downsample the input image by 16, thus the input sequence length for the transformer is $N = HW / 16^2$. 
The receptive field of ResNet-18, ResNet-34, and ResNet-50 are 435, 899, and 427, respectively~\cite{araujo2019computing}. 

\textbf{Linear projection.} 
Cordonnier, \etal~\cite{cordonnier2019relationship} initially introduced the concept of linear projection, 
and \ac{vit}~\cite{dosovitskiy2020image} further demonstrated its scalability via large-scale pretraining. 
Building upon the simplicity of this design, \ac{vilt}~\cite{kim2021vilt} modified the visual encoder by replacing the region supervision and the convolutional backbone with the linear projection. 
The modification significantly accelerated the inference speed, all while maintaining the model's expressive power. 
The linear projection layer reshapes the image $I \in \mathbb{R}^{H \times W \times C}$ into a sequence of flattened 2D patches $I_p \in \mathbb{R}^{N \times (P^2 \cdot C)}$, where $C$ is the number of channels, $(P, P)$ is the size of each image patch, and $N = HW/P^2$ is the number of patches, which is also the input sequence length for the transformer of the textual decoder. 
It is implemented by a stride $P$, kernel $P \times P$ convolution applied to the input image. 
The receptive field of the linear projection is the same as the patch size $P$. 
We denote ``LinearProj-28'' as a linear projection layer of $P = 28$.

\textbf{Convolutional stem.} 
The convolutional stem was first introduced to replace the linear projection in \acp{vit}, serving as the earliest stage of processing input image~\cite{xiao2021early}. 
This early convolution enhances optimization stability and improves peak performance and robustness~\cite{singh2023revisiting, peng2023robust}. 
To implement the convolutional stem, we employ a stack of stride 2, kernel $3\times 3$ convolutions, followed by a single stride 1, kernel $1 \times 1$ convolution at the end to match the $d$--dimension feature of the transformer.
We tune the receptive field of the convolutional stem by varying the kernel size, layers of convolutions, and input image size. 
Denote ``ConvStem'' as a visual encoder that uses convolutional stem as the \ac{cnn} backbone. 

\subsection{Textual Decoder} 
\label{sec:structure}
The table structure can be defined either using HTML tags~\cite{li2019tablebank} or LaTeX symbols~\cite{ye2021pingan}. 
Since these different representations are convertible, we select HTML tags as they are the most commonly used format in dataset annotations~\cite{zhong2020image, zheng2021global,li2019tablebank}. 
Our HTML structural corpus has 32 tokens, including
(1) starting tags \texttt{<thead>}, \texttt{<tbody>}, \texttt{<tr>}, \texttt{<td>}, along with their corresponding closing tags;
(2) spanning tags \texttt{<td}, \texttt{>} with the maximum values 
 for \texttt{rowspan} and \texttt{colspan} set at 10;
(3) special tokens \texttt{<sos>}, \texttt{<eos>}, \texttt{<pad>}, and \texttt{<unk>}. 
The textual decoder is a stack of transformer encoder and decoder layers, primarily comprised of multi-head attention and feed-forward layers.
During training, we apply the teacher forcing so that the transformer decoder receives ground truth tokens. 
At inference time, we employ greedy decoding, using previous predictions as input for the transformer decoder.

\subsection{Loss Function} 
\label{sec:loss}
We formulate the training loss based on the language modeling task because the HTML table tags are predicted in an autoregressive manner. 
Denote the probability of the $i$th step prediction $p(t_{s_i} | I, t_{s_1: s_{i - 1}}; \theta)$, we directly maximize the correct structure prediction by using the following formulation:
\begin{equation}
    \theta^\ast = \argmax_\theta \sum_{(I, T_s)} \log p(T_s | I; \theta),
\end{equation}
where $\theta$ are the parameters of our model, $I$ is a tabular image, and $T_s$ is the correct structure sequence.
According to language modeling, we apply the chain rule to model the joint probability over a sequence of length $n$ as 
\begin{equation}
    \log p(T_s | I; \theta) = \sum_{i=2}^n \log p(t_{s_i} | I, t_{s_1: s_{i - 1}}; \theta).
\end{equation}
The start token $t_{s_1}$ is a fixed token \texttt{<sos>} in both training and testing. 

\section{Experiments}
\label{sec:experiments}

\subsection{Settings}
\label{sec:implementation}

\textbf{Architecture.} 
The textual decoder has four layers of transformer decoders. 
For the visual encoder using the \ac{cnn} backbone, we employ two layers of transformer encoder as this is shown to be the optimal setting~\cite{nassar2022tableformer}.
For all other visual encoders, we use the same layers of transformer encoders and decoders~\cite{vaswani2017attention}.
All transformer layers have an input feature size of $d = 512$, a feed-forward network of 1024, and 8 attention heads. 
The maximum length for the HTML sequence decoder is set to 512.

\textbf{Training.} 
All models are trained with the AdamW optimizer~\cite{loshchilov2017decoupled} as transformers are sensitive to the choice of the optimizer. 
We employ a step learning rate scheduler, starting with an initial learning rate of 0.0001 for 12 epochs, which is then reduced by a factor of 10 for the subsequent 12 epochs. 
To prevent overfitting, we set the dropout rate to 0.5.
The input images are resized to $448 \times 448$ by default~\cite{zhong2020image, nassar2022tableformer}, and normalized using mean and standard deviation. 

\textbf{Dataset and metric.} 
We train and test on PubTabNet~\cite{zhong2020image} with $\sim$509k annotated tabular images, which is one of the largest publicly accessible \ac{tsr} datasets. 
PubTabNet uses \ac{teds} score~\cite{zhong2020image} as the evaluation metric.
It converts the HTML tags of a table into a tree structure and measures the edit distance between the prediction $T_{pred}$ and the groundtruth $T_{gt}$:
\begin{equation}
    \text{TEDS}(T_{pred}, T_{gt}) = 1 - \frac{\text{EditDist}(T_{pred}, T_{gt})}{\max (\lvert T_{pred} \rvert, \lvert T_{gt} \rvert)},
\end{equation}
A shorter edit distance indicates a higher degree of similarity, leading to a higher \ac{teds} score. 
Tables are classified as either ``simple'' if they do not contain row spans or column spans, or ``complex'' if they do.
We report the \ac{teds} scores for simple tables, complex tables, and the overall dataset.

\subsection{Quantitative Analysis}
\label{sec:quantitative}

\begin{table}[!htbp]
\small
\centering
\caption{
The convolutional stem (ConvStem) effectively bridges the performance gap between linear projection (LinearProj-28) and \ac{cnn} backbone (ResNet-18) while significantly reducing the model complexity of ResNet-18.
Differences are compared with ResNet-18.
In comparison to all the models, our ConvStem demonstrates comparable performance to the current \ac{sota} TableFormer with substantially lower model complexity.
}
\begin{tabular}{lrr|r@{\hspace{1mm}}lr@{\hspace{1mm}}lr@{\hspace{1mm}}l}
\toprule
    & & & \multicolumn{6}{c}{\acs{teds} (\%)} \\
    Model & \#Param. & \acs{mac} & \multicolumn{2}{c}{Simple} & \multicolumn{2}{c}{Complex} & \multicolumn{2}{c}{All} \\
\midrule
    ResNet-18 & 28.70M & 42.22G & 98.31 & & 94.50 & & 96.45 \\
    LinearProj-28 & 22.67M & 10.28G & 94.12 & \textcolor{myred}{ -4.19} & 86.62 & \textcolor{myred}{ -7.88} & 90.45 & \textcolor{myred}{ -6.00}\\
    \textbf{ConvStem}  & 24.08M & 22.36G & \textbf{98.33} & \textcolor{mygreen}{+0.02} & \textbf{94.66} & \textcolor{mygreen}{+0.16} & \textbf{96.53} & \textcolor{mygreen}{+0.08} \\
\midrule
    EDD~\cite{zhong2020image} & - & - & 91.1 & & 88.7 & & 89.90 \\
    GTE~\cite{zheng2021global} & - & - & - & & - & & 93.01 \\
    Davar-Lab~\cite{jimeno2021icdar} & - & - & 97.88 & & 94.78 & & 96.36 \\
    TableFormer~\cite{nassar2022tableformer} & >28.70M & >42.22G & 98.50 & & 95.00 & & 96.75 \\
\bottomrule
\end{tabular}
\label{tab:sota}
\end{table}

Table~\ref{tab:sota} demonstrates the effectiveness of the convolutional stem in comparison to \ac{cnn} backbone and linear projection by showing the results of all three types of visual encoders introduced in Sec.~\ref{sec:visual}. 
When comparing LinearProj-28 to the baseline ResNet-18, where the \ac{cnn} backbone is replaced by a linear projection, we observe a significant decrease in performance: $\sim 4$ \ac{pp} for simple tables, $\sim 8$ \ac{pp} for complex tables, and $\sim 6$ \ac{pp} in overall \ac{teds} score. 
Reintroducing 5 convolution layers by constructing the convolutional stem, ConvStem not only bridges the performance gap between the linear projection and the \ac{cnn} backbone but also slightly outperforms ResNet-18, especially for the complex table, highlighting the effectiveness of this design.
In terms of model complexity, we measure both the total number of parameters and \ac{mac}, both are computed by the ptflops library~\cite{ptflops}.
It is worth noting that ConvStem substantially reduces ResNet-18's model complexity, with the number of parameters reduced by 4.62M (28.70M $\rightarrow$ 24.08M) and \ac{mac} reduced by 19.86G (42.22G 
$\rightarrow$ 22.36G).

Next, we compare our models to \ac{sota} architectures.
However, computing the model complexity is not straightforward, as most literature has yet to release the code. 
Our ResNet-18 is similar to TableFormer~\cite{nassar2022tableformer} except that we omit the cell bounding box decoder as our paper focuses on the table structure. 
Therefore, the total complexity of TableFormer must be higher than that of our ResNet-18 implementation.
We hypothesize that the slight improvement in TableFormer is due to the additional \ac{iou} loss~\cite{rezatofighi2019generalized} used for training the cell bounding box decoder.  
In comparison to all the models, our ConvStem performs on par with the current \ac{sota} TableFormer with substantially lower model complexity.

\begin{figure}[t]
\centering
\includegraphics[width=0.9\linewidth]{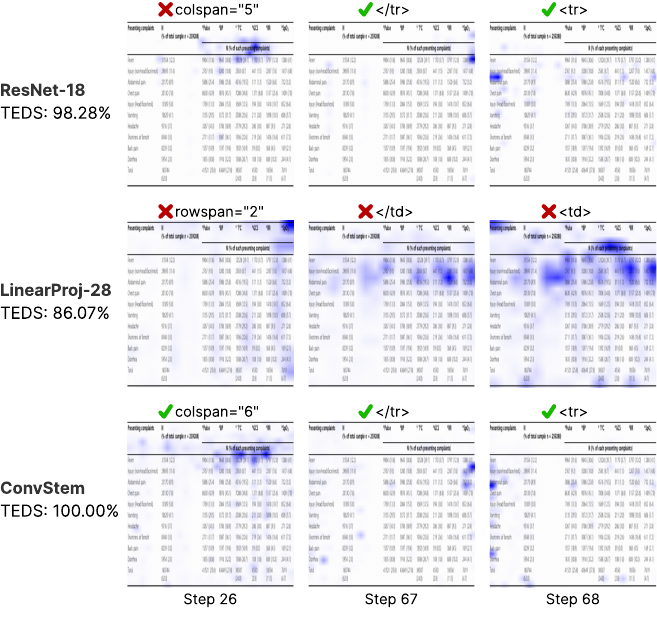}
\caption{
The convolutional stem accurately reconstructs the table; however, the \ac{cnn} backbone miscounts columns in the header spanning cell. 
The linear projection's scattered attention leads to the omission of predicting several complex table structures. 
The cross-attention maps (highlighted in blue) illustrate how the model processes various components of a table structure. 
We calculate the \ac{teds} score for each model on this complex table. 
The HTML prediction is displayed above each individual attention map. 
The first column shows a header data cell that spans six columns, while the second and third columns visualize the beginning of a new data row. 
}
\label{fig:attention}
\end{figure}

\subsection{Qualitative Analysis}
\label{sec:qualitative}

In Fig.~\ref{fig:attention}, we showcase the cross-attention maps of all three types of visual encoders, illustrating how the model processes various components of a table structure. 
In the first column of Fig.~\ref{fig:attention}, we visualize a header data cell that spans six columns. 
Both ResNet-18 and ConvStem accurately focus on this spanning cell, but ResNet-18 miscalculates the number of columns it spans as the attention of ConvStem is more evenly scattered across six columns. 
For LinearProj-28, the attention erroneously centers on the top-right data cell and predicts a multi-row structure, \ie, \texttt{rowspan=``2''}. 
In the second and third columns of Fig.~\ref{fig:attention}, we visualize the beginning of a new data row. 
For ResNet-18 and ConvStem, the attention concentrates on the end of the preceding row when predicting \texttt{</tr>} and successfully transitions to the start of the subsequent row when predicting the next token \texttt{<tr>}.
However, LinearProj-28's attention still focuses on the data cell due to the incorrect prediction in the early steps. 
In summary, the convolutional stem accurately reconstructs the table, whereas the \ac{cnn} backbone makes an error in calculating the number of columns in a spanning cell. 
The attention of the linear projection is often dispersed, leading to the omission of predicting several complex table structures.

\subsection{Ablations}
\label{sec:ablation}

This section investigates the root cause of the performance degradation from \ac{cnn} backbone to linear projection and explains why convolutional stem can bridge this performance gap.   
We conduct a series of ablation experiments to delve deeper into the impacts of the receptive field and sequence length of the transformer. 
Sec.~\ref{sec:concept} introduces the concept of two factors that affect performance: \ac{rf} ratio and sequence length $N$, and Sec.~\ref{sec:ve_analysis} analyzes these factors across different visual encoders, comparing their effects.

\begin{table}[!htbp]
\small
\centering
\caption{
The convolutional stem balances the model complexity, \ac{rf} ratio, and sequence length, while still achieving competitive performance.
We show ablations of how \ac{rf} ratio and sequence length $N$ impact the performance of different visual encoders.
In general, higher \ac{rf} ratio and longer $N$ benefit \ac{tsr}, especially with complex tables. 
Among these visual encoders, the \ac{cnn} backbone is performant with exhaustive \ac{rf} but exhibits high model complexity. 
Linear projection is the simplest but suffers in terms of performance due to limited \ac{rf} and sequence length. 
We highlight the best model for each type of visual encoder.
}
\begin{tabular}{lrr@{\hspace{2mm}}r@{\hspace{2mm}}r@{\hspace{2mm}}r@{\hspace{2mm}}r|c@{\hspace{3mm}}c@{\hspace{3mm}}c}
\toprule
    & & & & & & & \multicolumn{3}{c}{\acs{teds} (\%)} \\
    Model & \#Param. & \acs{mac} & \#Conv. & Kernel & \acs{rf} ratio (\%) & $N$ & Simple & Complex & All \\
\midrule
    ResNet-18 & 28.70M & 42.22G & 17 & - & 97.10 & 784 & 98.31 & 94.50 & 96.45 \\
    \textbf{ResNet-34} & 38.80M & 71.87G & 33 & - & 100.00 & 784 & \textbf{98.44} & \textbf{95.00} & \textbf{96.76} \\
    ResNet-50 & 41.81M & 79.60G & 49 & - & 95.31 & 784 & 98.44 & 94.89 & 96.70 \\
\midrule
    LinearProj-14 & 21.89M & 24.03G & 1 & 14 & 3.13 & 1024 & 91.15 & 83.11 & 87.22 \\
    LinearProj-16 & 21.86M & 19.21G & 1 & 16 & 3.57 & 784 & 91.69 & 83.25 & 87.56\\
    LinearProj-28 & 22.67M & 10.28G & 1 & 28 & 6.25 & 256 & 94.12 & 86.62 & 90.45 \\
    \textbf{LinearProj-56}  & 26.28M & 7.60G & 1 & 56 & 12.50 & 64 & \textbf{95.61} & \textbf{88.57} & \textbf{92.17}\\
    LinearProj-112 & 40.73M & 6.98G & 1 & 112 & 25.00 & 16 & 94.49 & 86.55 & 90.61\\
\midrule
    ConvStem-R1 & 22.53M & 20.69G & 5 & 3 & 6.92 & 784 & 97.68 & 93.36 & 95.57 \\
    ConvStem-R2 & 24.11M & 22.36G & 5 & 5 & 12.82 & 784 & 98.30 & 93.89 & 96.14 \\
    ConvStem-R3 & 22.14M & 18.67G & 4 & 5 & 12.95 & 729 & 97.97 & 93.99 & 96.02 \\
    \textbf{ConvStem}  & 24.08M & 22.36G & 5 & 5 & 13.62 & 729 & \textbf{98.33} & \textbf{94.66} & \textbf{96.53} \\
\midrule
    ConvStem-N1 & 22.39M & 10.55G & 5 & 3 & 12.30 & 256 & 96.99 & 91.53 & 94.32 \\
    ConvStem-N2 & 23.98M & 17.67G & 5 & 5 & 15.56 & 528 & 98.03 & 93.65 & 95.89\\
    \textbf{ConvStem}  & 24.08M & 22.36G & 5 & 5 & 13.62 & 729 & \textbf{98.33} & \textbf{94.66} & \textbf{96.53} \\
    ConvStem-N3 & 24.18M & 26.82G & 5 & 5 & 12.10 & 900 & 98.27 & 94.65 & 96.50 \\
\bottomrule
\end{tabular}
\label{tab:ablation}
\end{table}

\subsubsection{RF Ratio \& Sequence Length}
\label{sec:concept}
The \ac{rf} is a key parameter in understanding the degree to which input signals may impact output features, and mapping features at any part of the network to the region in the input that generates them~\cite{le2017receptive}.
We define \ac{rf} ratio as the size of this input region divided by the size of the entire input image on one side. 
Notably, the \ac{rf} is solely determined by the model's architecture, whereas increasing the image size alone has been demonstrated to enhance accuracy~\cite{liu2022convnet}. 
Therefore, the \ac{rf} ratio is defined to exclude the influence of image size.

The sequence length $N$ is another important factor influencing the performance of the transformer.
Training with a longer context generally yields higher-quality models~\cite{tay2020long}, but the bottlenecks lie in the computation cost and memory of the attention layer: 
doubling $N$ would quadruple the runtime and memory requirements~\cite{dao2022flashattention}.
In our \ac{tsr} model, the transformer receives the flattened feature map from the visual encoder, so $N$ is quadratic to the size of the feature map.
Thus, reducing the feature map size can significantly improve the training and inference speed. 

\subsubsection{Analysis of Three Visual Encoders}
\label{sec:ve_analysis}

In Table~\ref{tab:ablation}, we explore variations in the number of convolutional layers, kernel size, and input image size for each type of visual encoder. 
Alongside these configurations, we list details on total parameters, \ac{mac} and \ac{teds} score for each model. 

\textbf{CNN backbone.}
We test three \ac{cnn} backbones: ResNet-18, ResNet-34, and ResNet-50. 
Sequence length $N$ remains consistent across all three ResNets as they all downsample the input by a factor of 16. 
Comparing ResNet-18 to ResNet-34, the \ac{teds} increases along with the increase of the \ac{rf} ratio, as expected.
In contrast, when comparing ResNet-34 to ResNet-50, the \ac{teds} of simple tables are similar, but ResNet-50 has a worse \ac{teds} of complex tables, despite having 3.01M more parameters. 
This discrepancy is exactly due to the reduction in the \ac{rf} ratio in ResNet-50. 

\textbf{Linear projection.} 
The \ac{rf} ratio and $N$ are inversely correlated in linear projection. 
An increase in the \ac{rf} ratio means a larger patch size $P$, resulting in a shorter sequence length $N = HW / P^2$. 
We ablate on five different patch sizes, ranging from 14 to 112. 
With the increase of the patch size, we can clearly observe a boost in the \ac{teds} score, especially for the complex table. 
LinearProj-56 achieves the peak performance, and we observe a sharp decline in \ac{teds} score when the \ac{rf} ratio continues to increase. 
As previously mentioned, the \ac{rf} ratio and $N$ exhibit an inverse correlation, and a significantly low $N$ reduces cross-attention among different patches, which leads to a performance reduction 
Consequently, due to this correlation, the performance of linear projection is capped by a specific patch size. 

\textbf{Convolutional stem.}
In Table~\ref{tab:ablation}, we perform separate ablations on the convolutional stem by tuning the number of convolution layers, kernel size, and input image size. 
While keeping $N$ fixed at $27^2$ or $28^2$, we observe the consistent trend that the \ac{teds} increases along with the \ac{rf} ratio. 
When we constrain the \ac{rf} ratio within a certain range, increasing the value of $N$ further enhances performance.

In general, a higher \ac{rf} ratio and longer sequence length to transformers are beneficial for \ac{tsr}, particularly when dealing with complex tables. 
As shown in Fig.~\ref{fig:pipeline}, a small \ac{rf} provides only minimal structural information, while a medium-sized \ac{rf} offers sufficient context and distinguishable features. 
Among the three types of visual encoders, the \ac{cnn} backbone is performant with exhaustive \ac{rf} but exhibits high model complexity. 
Linear projection is the simplest but suffers in terms of performance due to limited \ac{rf} and sequence length. 
In contrast, the convolutional stem balances the model complexity, \ac{rf} ratio, and sequence length, while still achieving competitive performance. 
These results also demonstrate the benefits of injecting the inductive bias of early convolution, especially locality, into the learning ability of transformers.

\section{Conclusion}
\label{sec:conclusion}
In this work, we design a lightweight visual encoder for \ac{tsr} without sacrificing expressive power.
We discover that a convolutional stem can match classic \ac{cnn} backbone performance, with a much simpler model.
The convolutional stem strikes an optimal balance between two crucial factors for high-performance \ac{tsr}: a higher \ac{rf} ratio and a longer sequence length.
This allows it to ``see'' an appropriate portion of the table and ``store'' the complex table structure within sufficient context length for the subsequent transformer. 
We conducted reproducible ablation studies and open-sourced our code to enhance transparency, inspire innovations, and facilitate fair comparisons in our domain as tables are a promising modality for representation learning.

\clearpage

\bibliography{ref}
\bibliographystyle{unsrtnat}

\end{document}